\documentclass[preprints,article,accept,moreauthors,dvi2pdf]{Definitions/mdpi} 

\usepackage[english]{babel}
\usepackage[utf8]{inputenc}
\usepackage{amsmath} 
\usepackage{amssymb}
\usepackage{subcaption}
\usepackage{multirow}
\usepackage{capt-of}
\usepackage{gensymb}
\usepackage{siunitx}
\usepackage{xcolor}
\usepackage[mathscr]{euscript}
\usepackage[capitalise,nameinlink]{cleveref}
  \crefname{section}{Sect.}{Sect.}
  \Crefname{section}{Section}{Sections}
  \crefname{figure}{Fig.}{Fig.}
  \Crefname{figure}{Figure}{Figures}
  \crefname{table}{Tabl.}{Tabl.}
  \Crefname{table}{Table}{Tables}

\firstpage{1} 
\pubvolume{xx}
\issuenum{1}
\articlenumber{5}
\pubyear{2019}
\copyrightyear{2019}
\history{Received: date; Accepted: date; Published: date}

\Title{Variable Compliance Control for Robotic Peg-in-Hole Assembly: A Deep-Reinforcement-Learning Approach}

\Author{Beltran-Hernandez Cristian C. $^{1,*}$\orcidA{},
Damien Petit $^{1}$\orcidB{},\\
Ixchel G. Ramirez-Alpizar $^{2}$\orcidC{},
and Kensuke Harada $^{1,2}$\orcidD{}}

\AuthorNames{Beltran-Hernandez Cristian Camilo, Damien Petit, Ixchel G. Ramirez-Alpizar and Kensuke Harada}

\address{%
$^{1}$ \quad Graduate School of Engineering Science, Osaka University, Japan;\\
$^{2}$ \quad Automation Research Team, Industrial CPS Research Center, National Institute of Advanced Industrial Science and Technology (AIST), Japan
}

\corres{Correspondence: beltran [at] hlab.sys.es.osaka-u.ac.jp}

\abstract{
Industrial robot manipulators are playing a more significant role in modern manufacturing industries. Though peg-in-hole assembly is a common industrial task that has been extensively researched, safely solving complex high-precision assembly in an unstructured environment remains an open problem.
Reinforcement-learning (RL) methods have  proven to be successful in autonomously solving manipulation tasks. However, RL is still not widely adopted on real robotic systems because working with real hardware entails additional challenges, especially when using position-controlled manipulators. 
The main contribution of this work is a learning-based method to solve peg-in-hole tasks with hole-position uncertainty. We propose the use of an off-policy model-free reinforcement-learning method, and we bootstraped the training speed by using several transfer-learning techniques (sim2real) and domain randomization. Our proposed learning framework for position-controlled robots was extensively evaluated on contact-rich insertion tasks in a variety of environments.
}

\keyword{reinforcement learning; compliance control; robotic assembly; sim2real; domain randomization}

\featuredapplication{assembly tasks with industrial robot manipulators.}

\begin{document}


\section{Introduction}






    Autonomous robotic assembly is an essential component of industrial applications. Industrial robot manipulators are playing a more significant role in modern manufacturing industries with the goal of improving production efficiency and reducing costs. Though peg-in-hole assembly is a common industrial task that has been extensively researched, safely solving complex high-precision assembly in an unstructured environment remains an open problem \cite{kroemer2019review}.
    
    Most common industrial robots are joint-position-controlled. For this type of robot, compliance control is necessary to safely attempt contact-rich tasks, or the robot is prone to causing large unsafe assembly forces even with tiny position errors. Compliant robot assembly tasks have been studied in two ways, passive and active methods. In  passive methods, a mechanical device called remote center compliance (RCC) \cite{whitney1982quasi} is placed between the robot's wrist and gripper. The passive compliance provided by the RCC lets the gripper move perpendicularly to the peg’s axis and rotate freely so as to reduce resistance. However, the passive method does not work well with high-precision assembly \cite{tsuruoka19973d}. On the other hand, active compliant methods correct assembly errors through sensor feedback. In general, these methods use force sensors to detect the external forces and moments, and design control strategies  on the basis of  dynamic models of the task to minimize  contact force \cite{zhang2017force}. Some active methods mimic human compliance during assembly \cite{fukumoto2018force}. Nevertheless, most of these assembly methods are not practical to use in real applications. Model parameters need to be identified, and  controller gains need to be tuned. In both cases, the process is manually engineered for specific tasks, which requires a lot of time, effort, and expertise. These approaches  are also not robust to uncertainties and do not generalize well to variations in the environment.
    
    To reduce human involvement and increase  robustness to uncertainties, the most recent research has been focused on learning  assembly skills either from human demonstrations \cite{kyrarini2019robot} or directly from interactions with the environment \cite{sutton2018reinforcement}. The present research focuses on the latter.
    
    Reinforcement-learning (RL) methods allow for agents to learn complex behaviors through interactions with the surrounding environment, and by maximizing  rewards received from the environment; ideally, the agents’ behavior can generalize to unseen scenarios or tasks \cite{sutton2018reinforcement}. Therefore, RL can be applied to robotic agents to learn high-precision assembly skills instead of only transferring human skills to the robot program \cite{yang2018learning}. Recent studies showed the importance of RL for robotic manipulation tasks \cite{levine2018learning,pinto2016supersizing,gu2017deep}, but none of these methods can be applied directly to high-precision industrial applications due to the lack of fine motion control. 
    
    In \cite{nuttin1997learning}, an RL technique was used to learn a simple peg-in-hole insertion operation. Similarly, Inuo et al. \cite{inoue2017deep} proposed a robot skill-acquisition approach by training a recurrent neural network to learn a peg-in-hole assembly policy. However, these approaches used a finite number of actions by discretizing the action space, which has many limitations in continuous-action control tasks \cite{lillicrap2015continuous}, as is the case for robot control, which is continuous and high-dimensional. 
    
    Xu et al. \cite{xu2018feedback} proposed learning  dual peg insertion by using the deep deterministic policy gradient \cite{silver2014deterministic} (DDPG) algorithm with a fuzzy reward system. Similarly, Fan et al. \cite{fan2019learning} used DDPG combined with guided policy search (GPS) \cite{levine2013guided} to learn high-precision assembly tasks. Luo et al. \cite{luo2018deep} also used GPS to learn a peg-in-hole tasks on a deformable surface. Nevertheless, these methods learn policies that control the motion trajectory only while they require the manual tuning of  force control gains; therefore,  they do not scale well to variations of the environment. 
    
    Ren et al. \cite{ren2018learning} proposed the use of DDPG to simultaneously control position and force control gains, but they assumed the  geometric knowledge of the insertion task, which made the learned policies inflexible to be applied to different insertion tasks. To solve high-precision assembly tasks, our approach focused on learning policies that simultaneously control the robot’s motion trajectory and actively tune a compliant controller to unknown geometric constraints.
    
    Buchli et al. \cite{buchli2011learning} accomplished variable stiffness skill learning on robot manipulators by using an RL algorithm call-policy improvement with path integrals (PI2). However, the method was formulated for torque-control robots. Another similar approach was to use a flexible robot so as to focus only on the motion trajectory, as in \cite{lee2019making}; however, rigid position-controlled robots are still more widely used. Therefore, we focus on industrial robot manipulators, which are mainly position-based-controlled.
    
    \color{black}
    Abu- Dakka et al. \cite{abu2015adaptation} proposed a learning method based on iterative learning control (ILC). Their method is focus on transferring manipulation skills from demonstrations that provide a reference trajectory and force profile. In this work, we present a method that can learn manipulation skills without prior knowledge of a reference trajectory or force profile. However, our method supports the use of such prior knowledge to speed up the learning phase.
    
    The main contribution of this work is a robust learning-based framework for robotic peg-in-hole assembly given an uncertain goal position. Our method enables a position-controlled industrial robot manipulator to safely learn contact-rich manipulation tasks by controlling the nominal trajectory and, at the same time, learning variable force control gains for each phase of the task.
    We built this on the basis of our previous work \cite{beltranhern2020learning}. More specifically, the contributions of this work are:
    \begin{itemize}
        \item A robust policy representation based on time convolutional neural networks (TCN).
        \item Faster learning of control policies via domain transfer-learning techniques (sim2real) to greatly improve the training efficiency on real robots.
        \item Improved generalization capabilities of the learned control policies via domain randomization during the training phase on simulation. Although the effects of domain randomization have been researched \cite{chebotar2019closing,andrychowicz2020learning}, to the best of our knowledge, we are the first to study the effects of sim2real with domain randomization on contact-rich real-robot applications with position-controlled robots.
    \end{itemize}  
    The effectiveness of the proposed method is shown through extensive evaluation with a real robotic system on a variety of contact-rich peg-in-hole insertion tasks. 
    \color{black}

 \subsection{Problem Statement}\label{sec:problem-statement}
 
\begin{figure}[ht]
    \centering
    \includegraphics[height=0.35\textwidth]{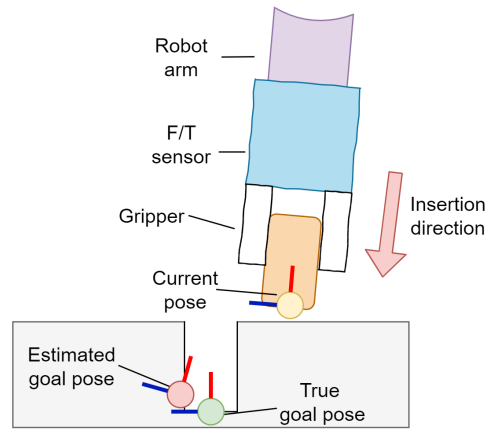}
    \caption{Insertion task with uncertain goal position.}
    \label{fig:insertion-environment}
\end{figure}

    In the present study, we considered a peg-in-hole assembly task that required the mating of two components. One of the components was grasped and manipulated by the robot manipulator, while the second component had a fixed position either via fixtures to an environment surface  or by being held by a second robot manipulator. \Cref{fig:insertion-environment} provides a 2D representation of the  considered insertion tasks  and the components assumed to be available to solve the task. The proposed method was designed for a position-controlled robot manipulator with a force/torque sensor at its wrist. Typically, these insertion tasks can be broadly divided into two main phases \cite{sharma2013intelligent}, search and insertion. During the search phase, the robot aligns the peg within the clearance region of the hole. In the beginning, the peg is located at a distance from the center of the hole in a random direction. The distance from the hole is assumed to be the “positional error”.  During the insertion phase, the robot adjusts the orientation of the peg with respect to the hole orientation, and pushes the peg to the desired position. We focused on both phases of the assembly task with the following assumptions:
    \begin{itemize}
        \item The manipulated object was already firmly grasped. However, slight changes of  object orientation within the gripper were possible during  manipulation.
        \item There was access to  imperfect prediction of the target end-effector pose (as shown in \Cref{fig:insertion-environment}) or a reference trajectory and its degree of uncertainty.
        \item The manipulated object was inserted in a direction parallel to the gripper's orientation.
    \end{itemize}
    We considered the second assumption fair given the advances in vision-recognition techniques, where the 6D pose of objects could be estimated from single RGB images \cite{zakharov2019dpod,peng2019pvnet} or  RGB images with depth maps (RGB-D) \cite{xiang2017posecnn,hodan2017t}. The high accuracy of the predictions are in many cases enough for robot manipulation. Moreover, this second assumption  included the specific case of using an assembly planner \cite{harada2018tool,masehian2020asppr}, where even if the initial position of the objects is known, the inevitable error throughout the manipulation (e.g. pick-and-place, grasping, and regrasping) that makes the position/orientation of the manipulated objects uncertain during the insertion phase. A reference trajectory could be similarly obtained from demonstrations \cite{nair2018overcoming,gupta2019relay,yan2020motionplan} when a complex motion is required to achieve the insertion. The last assumption allowed for defining a desired insertion force that may vary for different insertion tasks without loss of generalization.

\section{Materials and Methods}

\subsection{System Overview}\label{subsec:system-overview}

\begin{figure}[ht]
    \centering
    \includegraphics[width=\textwidth]{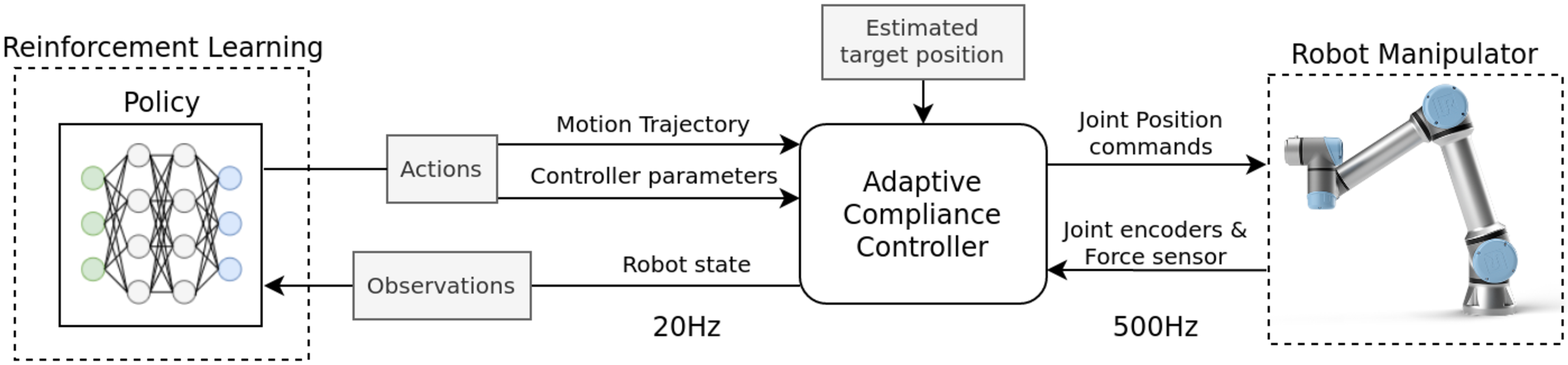}
    \caption{Our proposed framework. On the basis of estimated target position for an insertion task, our system learns a control policy that defines  motion-trajectory and force-control parameters of an adaptive compliance controller to control an industrial robot manipulator. }
    \label{fig:system-overview}
\end{figure}

    Our proposed system aims to solve assembly tasks with an uncertain goal pose.   \Cref{fig:system-overview} shows the overall system architecture. There were two control loops. The inner loop was an adaptive compliance controller; we chose to use a parallel position-force controller that was proven to work well for this kind of contact-rich manipulation tasks \cite{beltranhern2020learning}. The inner loop ran at a control frequency of 500 Hz, which is the maximum available in  Universal Robots e-series robotic arms\footnote{Robot details at https://www.universal-robots.com/e-series/}. Details of the parallel controller are provided in \Cref{subsec:parallel-control}. The outer loop was an RL control policy running at 20 Hz that provided subgoal positions and the parameters of the compliance controller. The outer loop's slower control frequency allowed for the policy to process the robot state and compute the next action to be taken by the manipulator, while the inner loop's precise high-frequency control would seek to achieve and maintain the subgoal provided by the policy. Details of the RL algorithm and the policy architecture are provided in \Cref{subsec:rl-algo}. Lastly, the input to the system was  estimated target position and orientation for the insertion task.

  Motion commands $\textbf{x}_c$ sent to the adaptive compliance controller corresponded to the pose of the robot's end effector. The pose was of the form $\textbf{x} = [\textbf{p},\phi]$, where $\textbf{p} \in \mathbb{R}^3 $ is the position vector, and $\phi \in \mathbb{R}^4$ is the orientation vector. The orientation vector was described using Euler parameters (unit quaternions), denoted as $\phi = \{\eta, \varepsilon\}$, where $\eta \in \mathbb{R}$ is the scalar part of the quaternion and $\varepsilon \in \mathbb{R}^3$ the vector part.
    
\subsection{Learning Adaptive-Compliance Control}\label{subsec:rl-algo}
\subsubsection{Reinforcement-Learning Algorithm}\label{subsubsec:rl-algo}
     Robotic reinforcement learning is a control problem where a robot, the agent, acts in a stochastic environment by sequentially choosing actions over a sequence of time steps. The goal is to maximize  a cumulative reward. Said problem was modeled as a Markov decision process. The environment is described by a state $\textbf{s} \in \mathscr{S}$. The agent can perform actions $\textbf{a} \in A$, and perceives the environment through observations $\textbf{o} \in O$ that may or not be equal to $\textbf{s}$. We considered an episodic interaction of finite time steps with a limit of $T$ time steps per episode. The agent's goal is to find a policy $\pi(\textbf{a}(t) \,|\, \textbf{o}(t))$ that selects actions $\textbf{a}(t)$ conditioned on  observations $\textbf{o}(t)$ to control the dynamical system. Given  stochastic dynamics $p(\textbf{s}(t+1) \,|\, \textbf{s}(t), \textbf{a}(t))$ and reward function $r(\textbf{s}, \textbf{a})$, the aim is to find a policy $\pi*$ that maximizes the expected sum of future rewards given by $R(t)=\sum_i^{\infty} \gamma r(s(t),a(t))$, with $\gamma$ being a discount factor \cite{sutton2018reinforcement}.

    In this work, we used an RL algorithm called soft actor critic (SAC), which is one of the state-of-the-art algorithms with high sample efficiency, ideal for real robotic applications. SAC \cite{Haarnoja2018SoftAO} is an off-policy actor-critic deep RL algorithm based on  maximal entropy. SAC aims to maximize the expected reward while also optimizing maximal entropy. The SAC agent optimizes a maximal-entropy objective, which encourages exploration according to a temperature parameter $\alpha$. The core idea of this method is to succeed at the task while acting as randomly as possible. Since SAC is an off-policy algorithm, it uses a replay buffer to reuse information from recent rollouts for sample-efficient training. Additionally, we used the distributed prioritized experience replay approach for further improvement \cite{horgan2018distributed}. Our implementation of the SAC algorithm was based on the TF2RL repository\footnote{TF2RL: Deep-reinforcement-learning library using TensorFlow 2.0. https://github.com/keiohta/tf2rl}.

\subsubsection{Multimodal Policy Architecture}\label{subsec:policy-arch}

\begin{figure}[ht]
    \centering
    \includegraphics[width=0.7\textwidth]{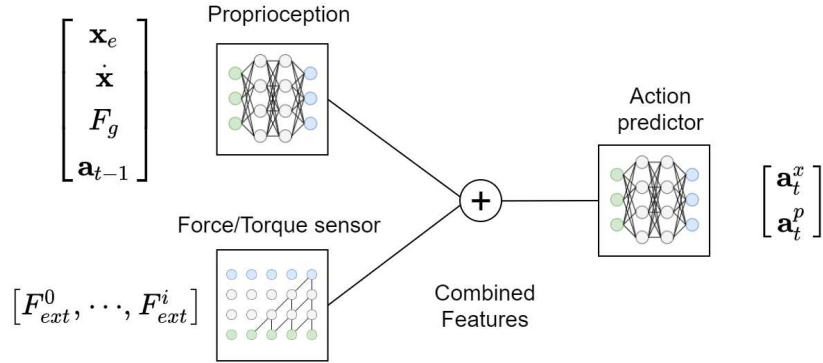}
    \caption{Control policy consisting of three networks. First,  proprioception information is processed through a 2-layer neural network. Second,  force/torque information is processed with a temporal convolutional network. Lastly, extracted features  from  first two networks are concatenated and processed on a 2-layer neural network to predict  actions.}
    \label{fig:policy-arch}
\end{figure}

    The control policy was represented using neural networks, as shown in \Cref{fig:policy-arch}. The policy input was the robot state. The robot state included the proprioception information of the manipulator and haptic information. Proprioception included the pose error between the current robot's end-effector position and  predicted target pose $\textbf{x}_e$, end-effector velocity $\Dot{\textbf{x}}$,  desired insertion force $F_g$, and  actions taken in the previous time step $\textbf{a}_{t-1}$. Proprioception feedback was encoded with a neural network with 2 fully connected layers with  activation function RELU to produce a 32-dimensional feature vector. For  force-torque feedback, we considered the last 12 readings from the six-axis F/T sensor, filtered using a low-pass filter, as a 12 x 6 time series: 
    \begin{equation}
        [F_{ext}^0,\,\dotsc,\, F_{ext}^{12}],  \quad\text{where}\quad F_{ext}^i = [F_x, F_y, F_z, M_x, M_y, M_z]
    \end{equation}
    The F/T time series was fed to a temporal convolutional network (TCN) \cite{bai2018empirical} to produce another 32-dimensional feature vector. The feature vectors from proprioception and haptic information were concatenated to obtain a 64-dimensional feature vector, and then fed to two fully connected layers to predict the next action.

     The policy outputs  actions for a parallel position-force controller. The policy produces two type of actions, $\textbf{a} \doteq [\textbf{a}_x, \textbf{a}_p]$,  where $\textbf{a}_x = [\textbf{p}, \phi]$ are position/orientation subgoals, and $\textbf{a}_p$ are parameters of the parallel controller. The specific parameters controlled by $\textbf{a}_p$ are described in \Cref{subsec:parallel-control}. 
  
\subsubsection{Compliance Control in Task Space}\label{subsec:parallel-control}
\begin{figure}[ht]
    \centering
    \includegraphics[width=0.65\textwidth]{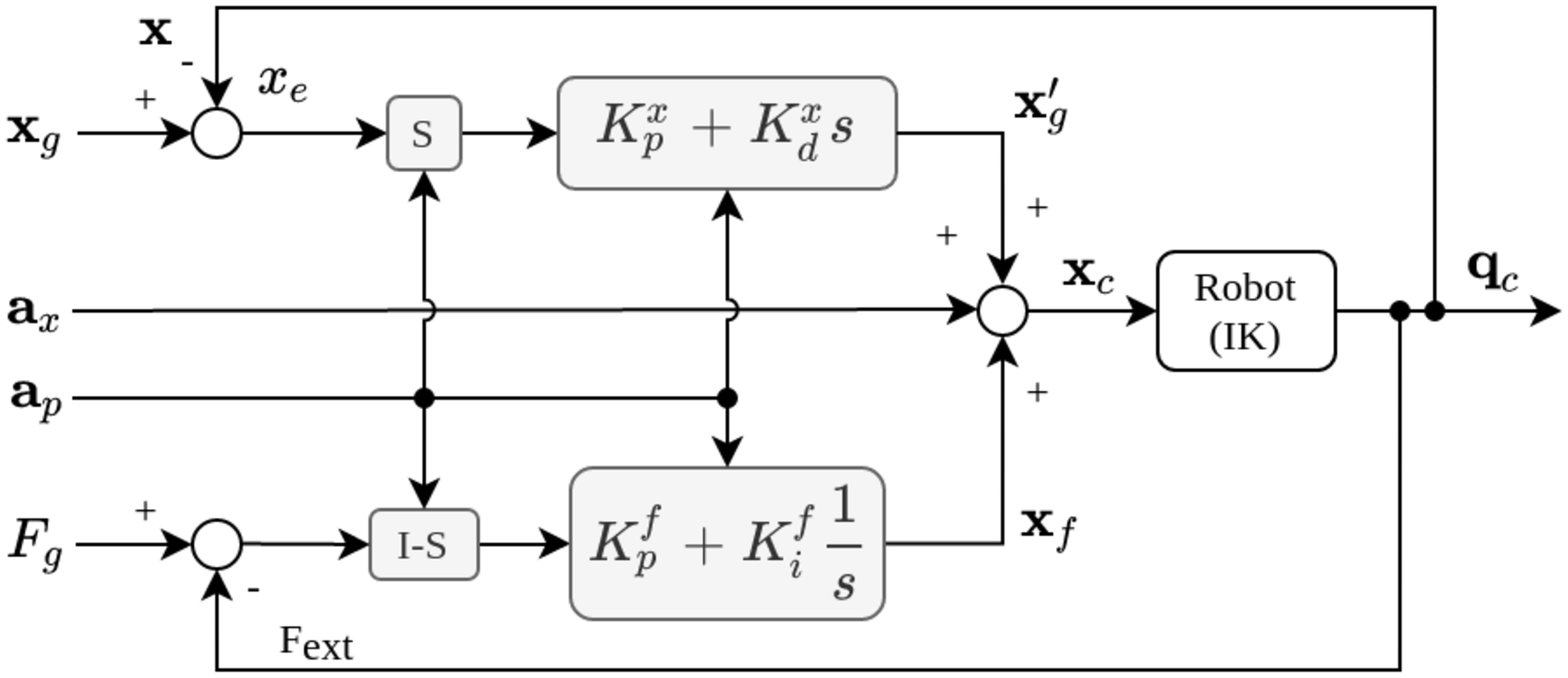}
    \caption{Adaptive parallel position-force control scheme \cite{beltranhern2020learning}. Inputs are the estimated goal position,  policy actions, and a desired contact force. Controller outputs the joint position commands for the robotic arm.}
    \label{fig:parallel-control}
\end{figure}

    Our proposed method uses a common force-control scheme combined with a reinforcement-learning policy to learn contact-rich manipulations with a rigid position-controlled robot.
    For the family of contact-rich manipulation tasks that require some sort of insertion, the parallel position-force control \cite{chiaverini1993parallel} performs better and can be learned faster than using an admittance control scheme when combined with an RL policy \cite{beltranhern2020learning}.
    
    The implemented parallel controller is depicted in \Cref{fig:parallel-control}. A PID parallel position-force control was used with the addition of a selection matrix to define the degree of the control of position and force over each direction. The control law consisted of a PD action on position, a PI action on force, a selection matrix, and  policy position action $\textbf{a}_x$,
    \begin{equation}
    \begin{aligned}
        \textbf{x}_c ~=  S (K_p^x\textbf{x}_e &+ K_d^x\Dot{\textbf{x}_e}) + \textbf{a}_x + (I-S)(K_p^fF_{e}+K_i^f\int F_{e} dt),
    \end{aligned}
    \end{equation}
    where $F_{e} = F_g - F_{ext}$, and $\textbf{x}_x$ is the commanded positions to the robot. The selection matrix is \[S = diag(s_1, ..., s_6),\quad s_j \in [0,1]\]
    
 where  values correspond to the degree of control that each controller has over a given direction.

    Our parallel control scheme had a total of 30 parameters, 12 from the position PD controller's gains, 12 from the force PI controller's gains, and 6 from  selection matrix $S$. We reduced the number of controllable parameters to prevent unstable behavior and to reduce  system complexity. For the PD controller, only  proportional gain $K_p^x$ was controllable, while  derivative gain $K_d^x$ was computed  on the basis of  $K_p^x$. $K_d^x$ was set to have a critically damped relationship as

    \[ K_d^x = 2\sqrt{K_p^x} \]
    Similarly, for the PI controller, only  proportional gain $K_p^f$ was controllable, and integral gain $K_i^f$ was computed with respect to $K_p^f$. In our experiments, $K_i^f$ was empirically set  to be $1\%$ of $K_p^f$. In total, 18 parameters were controllable. 
    In summary, the policy actions regarding the parallel controller's parameters are
    \[\textbf{a}_p = [K_p^x, K_p^f, S],  \quad \textbf{a}_p \in \mathbb{R}^{18}\]

    As a safety measure, we narrowed the agent choices for the force-control parameters by imposing upper and lower limits to each parameter. assuming we had access to some baseline gain values $P_{\text{base}}$. We defined a range of potential values for each parameter as $[P_{\text{base}}-P_{\text{range}}, P_{\text{base}}+P_{\text{range}}]$ with  constant $P_{\text{range}}$ defining the size of the range. We mapped policy actions $\textbf{a}_p$ from  range $[-1, 1]$ to each parameter's range. $P_{\text{base}}$ and $P_{\text{range}}$ are the hyperparameters of our method.

\subsection{Task's reward function}\label{subsubsec:cost-function}
    For all considered insertion tasks, the same reward function was used:
    \begin{equation}
    \begin{aligned}
        {r}(\textbf{s},\textbf{a}) = w_{1}L_m(\|(F_{ext}-F_g)/F_{max}\|_2) + w_2\kappa,
    \end{aligned}
    \label{eq:reward-function}
    \end{equation}
    where $F_g$ is the desired insertion force, $F_{ext}$ is the contact force, and $F_{max}$ is the defined allowed maximal contact force. $L_m(y) = y \mapsto x, x \in [1,0]$ is a linear mapping in the range 1 to 0; thus, the closer to the goal and the lower the contact force, the higher the obtained reward. $||\cdot||_{1,2}$ is an L1,2 norm based on \cite{levine2018learning}. $\kappa$ is a reward defined as follows:
    \begin{equation}
    \kappa = \left\{\begin{matrix}
     100 + ((1-t/T) * 100), & \textrm{Task completed}\\ 
     -50,     & \textrm{Collision} \\
       0,     & \textrm{Otherwise}
    \end{matrix}\right.
    \label{eq:safety-reward}
    \end{equation}
    During training, the task was considered completed if the Euclidean distance between the robot's end-effector position and the true goal position was less than 1 mm. The agent was encouraged to complete the task as quickly as possible by providing an extra reward for every unused time step with respect to the maximal number of time steps per episode $T$. Moreover, we imposed a collision constraint where the agent was penalized for colliding with the environment by giving it a negative reward and by finishing the episode early. This collision constraint encourages safer exploration, as shown in our previous work \cite{beltranhern2020learning}. We defined a collision as exceeding  force limit $F_{max}$. Therefore, a collision detector  and  geometric knowledge of the environment were not necessary. Lastly, each component was weighted via $w$; all $w$s were hyperparameters. 

\subsection{Speeding Up Learning}
    Two strategies were adopted to speed up the learning process. First, the exploitation of prior knowledge using the idea of residual reinforcement learning. Second, we used a physics simulator to train the robot on a peg-insertion task and transfer the learned policy directly to the real robot (sim2real). 
    
    \subsubsection{Residual Reinforcement Learning}\label{subsubsec:residual-rl}
    To speed up the learning of the control policy for insertion tasks that require complex manipulation, we used residual reinforcement learning \cite{Johannink2019residualRL,Silver2018ResidualPL}. The goal is to leverage the training process by exploiting prior knowledge. With the assumption of an estimated target position or a reference trajectory, we could manually define a controller $\textbf{x}_g$. Then, said controller's signal would be combined with  policy action $\textbf{a}_x$. The objective was to avoid training the policy from scratch, and avoid the exploration of the entire parameter space. The position command sent to the robot was
    \begin{equation}
    \textbf{x}_c = (\textbf{x}'_{g} + \textbf{x}_f) + \textbf{a}_x,
    \end{equation}
    where $\textbf{x}'_{g}$ is the reference trajectory process through a PD controller, $\textbf{a}_x$ is the policy signal on  the position, and $\textbf{x}_f$ is the response to the contact force, as shown in \Cref{fig:parallel-control}. The first two terms came from the parallel controller. Therefore, the policy would just need to learn to adjust the reference trajectory to achieve the task.
    
    \subsubsection{Sim2real}\label{subsubsec:sim2real}
    The proposed method works on the robot's end-effector Cartesian task-space, which makes it easier to transfer learning from simulation to the real robot or even between robots \cite{bellegarda2019training}. For most insertion tasks, a simple peg-insertion task was used for training on a physics simulator. We used  simulator Gazebo 9 \cite{koenig2004design}. To close the reality gap between the physics simulator and  real-world dynamics, we used domain randomization \cite{tobin2017domain}. During training on the simulator, the following aspects were randomized:
    \begin{itemize}
        \item Initial/goal end-effector position: having random initial/goal positions helps the RL algorithm to find policies that generalize to a wide range of initial-position conditions.
        \item Object-surface stiffness: The RL agent also needs to learn to fine-tune the force-controller parameters to obtain a proper response to the contact force. Therefore, randomizing the stiffness of the manipulated objects helps it find policies that adapt to different dynamic conditions.
        \item Uncertainty error of  goal pose prediction: On a real robot, the prediction of the target pose comes from noisy sensory information, either from a vision-detection system or from known prior  manipulations (grasp and regrasp). Thus, during training on the simulation, we emulated this error by using normal Gaussian distribution with mean zero and standard deviation of a maximal distance error (for position and orientation).
        \item Desired insertion force: For different insertion tasks, a specific contact force is necessary for  insertion to succeed. As we considered  insertion force an input to the policy, during training, we randomized this value for each episode.
    \end{itemize}

\section{Experiments and results}

\subsection{Experiment Setup}
\begin{figure}[ht]
    \centering
    \includegraphics[width=.3\textwidth]{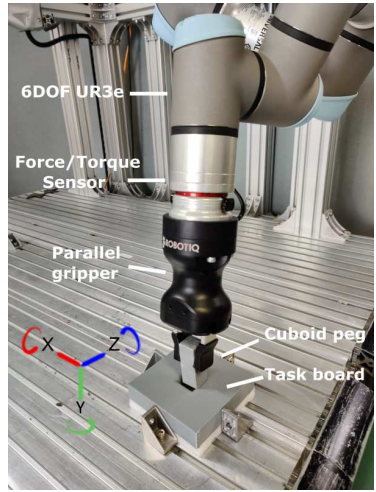}
    \caption{Real experiment environment with a 6-degree-of-freedom UR3e robotic arm. Cuboid peg and task board hole had a nonsmooth surface with  1.0 mm  clearance.}
    \label{fig:real-system}
\end{figure}

    Experimental validation was performed on a simulated environment using  Gazebo simulator \cite{koenig2004design} version 9, and on real hardware using a Universal Robot 3 e-series with a control frequency of up to 500 Hz. The robotic arm had a force/torque sensor mounted at its end effector, and a Robotiq Hand-e parallel gripper. In both environments, training of the RL agent was performed on a computer with an Intel i9-9900k CPU and  Nvidia RTX-2800 Super GPU. To control the robot agent, we used the Robot Operating System (ROS) \cite{quigley2009ros} with the Universal Robot ROS Driver\footnote{ROS driver for Universal Robot robotic arms developed in collaboration between Universal Robots and the FZI Research Center for Information Technology https://github.com/UniversalRobots/Universal\_Robots\_ROS\_Driver}. The experiment environment on the real robot is shown in \Cref{fig:real-system}.
    
\subsection{Training}
    During the training phase, the agent's task was to insert a cuboid peg into a task board on the simulated environment. The agent was trained for $500,000$ time steps, which, on average, takes about 5 hours to complete. During training, the environment was modified after each episode by randomizing one or several of the training conditions mentioned in \Cref{subsubsec:sim2real}. The range of values used for the randomization of the training conditions is shown in \Cref{table:training-conds}. The random goal position was selected from a defined set of possible insertion planes, as depicted in \Cref{fig:multi-goals}.
    
\begin{table}[ht]
\begin{minipage}[b]{0.45\linewidth}
    \centering
    \includegraphics[width=\textwidth]{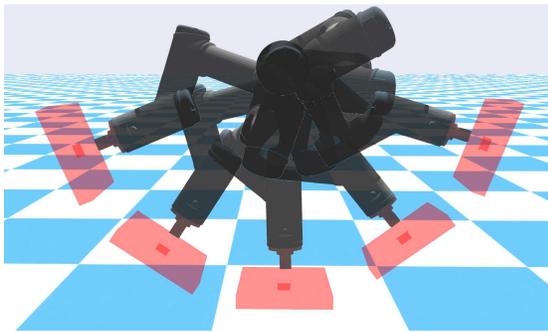}
    \captionof{figure}{Simulation environment. Overlay of randomizable goal positions.}
    \label{fig:multi-goals}
\end{minipage}
\begin{minipage}[b]{.55\linewidth}
    \begin{tabular}{|c|c|c|}
    \hline
    \multicolumn{2}{|c|}{\textbf{Condition}} & \textbf{Value range} \\ \hline
    \multirow{2}{*}{\begin{tabular}[c]{@{}c@{}}Initial position \\ (relative to goal)\end{tabular}} & Position (mm) & {[}-400, 400{]} \\ \cline{2-3} 
     & Orientation (\degree) & {[}-10, 10{]} \\ \hline
    \multirow{2}{*}{Uncertainty error} & Position (mm) & {[}-2, 2{]} \\ \cline{2-3} 
     & Orientation (\degree) & {[}-5, 5{]} \\ \hline
    \multicolumn{2}{|c|}{Desire insertion force (N)} & {[}0, 10{]} \\ \hline
    \multicolumn{2}{|c|}{\begin{tabular}[c]{@{}c@{}}Stiffness\\ (in Gazebo: \textit{surface/friction/ode/kp})\end{tabular}} & {[}\num{7.0e-4}, \num{1.0e-5}{]} \\ \hline
    \end{tabular}
    \caption{Randomized training conditions.}
    \label{table:training-conds}
\end{minipage}
\end{table}
    
    After training on the simulation, the learned policy was refined by retraining on the real robot for 3\% off the simulation time steps, which took about 20 minutes, to further account for the reality gap between  simulated  and  real-world physics dynamics.

\subsection{Evaluation}
\begin{figure}[ht]
    \centering
    \includegraphics[width=\textwidth]{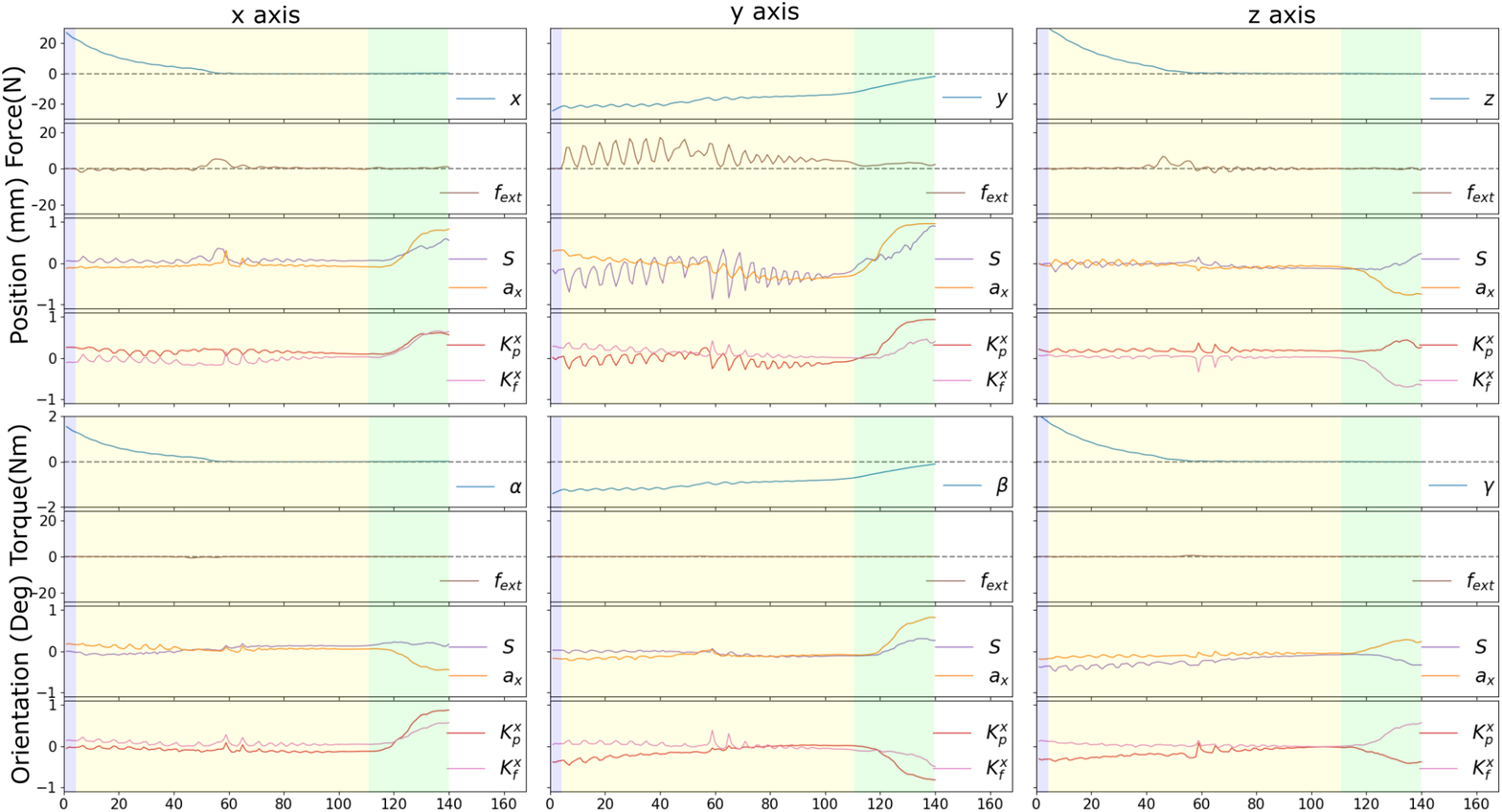}
    \caption{Performance of  learned policy (sim2real + retrain) on  3D-printed cuboid-peg-insertion task. Insertion direction was aligned with  y axis of  robot coordinate system. Relative distance from  robot's end effector to  goal position and  contact force is shown. The 24 policy actions besides the corresponding axis are also shown.}
    \label{fig:policy-perf}
\end{figure}

    The learned policy was initially evaluated on the real robot with a 3D-printed version of the cuboid peg in the hole-insertion task with the true goal pose. During  evaluation, observations and actions were recorded. \Cref{fig:policy-perf} shows the performance of the learned policy (sim2real + retrain). The figure shows the relative position of the end effector with respect to the goal position, the contact force, and the actions taken by the policy for each Cartesian direction normalized to the range of [-1, 1], as described in \Cref{subsec:parallel-control}.
    As shown in \cref{fig:insertion-environment}, the insertion direction was aligned with the y axis of the robot's coordinate system. In \Cref{fig:policy-perf}, we highlighted three phases of the task. Blue corresponds to the search phase in free space before contact with the surface, yellow is the search phase after initial contact with the environment, and green corresponds to the insertion phase. During the search phase, and particularly on the insertion direction (y axis), we could clearly observe that the learned policy properly reacted to  contact with the environment by quickly adjusting the force control parameters. On top of that, during the insertion phase, the learned policy changed its strategy from just minimizing contact force to a mostly position-control strategy to complete  insertion. This behavior is proper for this particular insertion task, as there is little resistance during the insertion phase, but it is not the desired behavior for other insertion tasks, as we discuss later in \Cref{subsubsec:varying-tasks}.

    Additionally, we compared the performance of the learned policy as a combination of sim2real and refinement on the real robot versus just learning on the real robot or just directly transferring the learned policy from the simulation (sim2real) without further training. We evaluated these policies on a 3D-printed version of the cuboid-peg-insertion task. Policies were tested 20 times with a random initial position assuming a perfect estimation of the goal position (true goal). \Cref{tab:accuracy-basic} shows the results of the evaluation. The three policies had a very high success rate, but the policy transfer from the simulation had difficulty with the real-world physics dynamics. As expected, the policy retrained from the simulation gave the best overall performance time.

\begin{table}[ht]
\centering
\begin{tabular}{|c|c|c|c|}
\hline
\textbf{Method}   & \textbf{Success Rate} & \textbf{Avg. Time Steps} & \textbf{Avg. Time (sec)} \\ \hline
\textbf{Scratch}  & 100\%                 & 109.6          & 5.48                \\ \hline
\textbf{Sim2real} & 95\%                  & 75.3           & 3.77                \\ \hline
\textbf{Ours}     & 100\%                 & 65.6           & 3.28                \\ \hline
\end{tabular}
\caption{Comparison of learning from scratch, straightforward sim2real, and sim2real + retraining (Ours). Test performed on a 3D printed cuboid peg insertion task assuming knowledge of the true goal position.}
\label{tab:accuracy-basic}
\end{table}

\subsection{Generalization}
    Now, to evaluate the generalization capabilities of our proposed learning framework, we use a series of environments with varying conditions.
    
    \subsubsection*{\textit{Varying degrees of Uncertainty error}}
    First, the learned policies are evaluated on the 3D printed cuboid peg insertion task where there is a degree of error on the estimation of the goal position. To clearly compare the performance of the different methods with different degrees of estimation error, we added and offset of position or orientation about the x-axis of the true goal pose. Nevertheless, for completeness we also evaluate the policies on goal poses with added random offset of translation, $[-1,\, 1]$ millimeters, and orientation, $[-5\degree,5\degree]$, on all directions. On each case, the policies were tested 20 times from random initial positions. Results are shown in \Cref{tab:acc-uncertainty}.
    
\begin{table}[ht]
\begin{tabular}{|c|c|c|c|c|c|c|c|c|c|c|c|}
\hline
\multicolumn{12}{|c|}{\textbf{Estimation error / Success rate}} \\ \hline
\textbf{} & \multicolumn{5}{c|}{\textbf{Position}} & \multicolumn{5}{c|}{\textbf{Orientation}} & \textbf{} \\ \hline
\textbf{Method} & \textbf{1 mm} & \textbf{2 mm} & \textbf{3 mm} & \textbf{4 mm} & \textbf{5 mm} & \textbf{1\degree} & \textbf{2\degree} & \textbf{3\degree} & \textbf{4\degree} & \textbf{5\degree} & \textbf{Random} \\ \hline
\textbf{Scratch} & 90\% & 90\% & 70\% & 55\% & 35\% & 100\% & 90\% & 80\% & 80\% & 50\% & 80\% \\ \hline
\textbf{Sim2real} & 90\% & 85\% & 75\% & 60\% & 40\% & 100\% & 90\% & 80\% & 80\% & 30\% & 75\% \\ \hline
\textbf{Ours} & \textbf{100\%} & \textbf{100\%} & \textbf{95\%} & \textbf{65\%} & \textbf{60\%} & \textbf{100\%} & \textbf{100\%} & \textbf{100\%} & \textbf{100\%} & \textbf{100\%} & \textbf{90\%} \\ \hline
\end{tabular}
\caption{Comparison of learning from scratch, straightforward sim2real and sim2real + retraining (Ours) with different degrees of goal-position uncertainty error. Test performed during  3D-printed cuboid-peg insertion task.}
\label{tab:acc-uncertainty}
\end{table}

    In all cases, the policy learned from the simulation with domain randomization and fine-tuned on the real robot gave the best results. If the difference between the physics dynamics on the simulation and the real world was too big, learning from scratch could yield better results than only transferring the policy from the simulation, as can be seen when the uncertainty error on orientation was too big (5\degree); where the friction with the environment makes the task much harder, such contact dynamics are difficult to simulate.

    \subsubsection*{\textit{Varying Environment Stiffness}}
    Second, the learned policy was also evaluated on different stiffness environments. \Cref{fig:env-stiff} shows the 3 environments considered for evaluation. High stiffness was the default environment. Medium stiffness was achieved by using a rubber band to hold the cuboid peg between the gripper fingers, adding a degree of static compliance. In addition to that, for the low-stiffness environment, a soft foam surface was added to further decrease  stiffness. The policies were evaluated from 20 different initial positions,  results are reported in \Cref{tab:stiffness-accuracy}.

\begin{table}[ht]
\begin{minipage}[b]{0.5\linewidth}
\centering
\includegraphics[width=\textwidth]{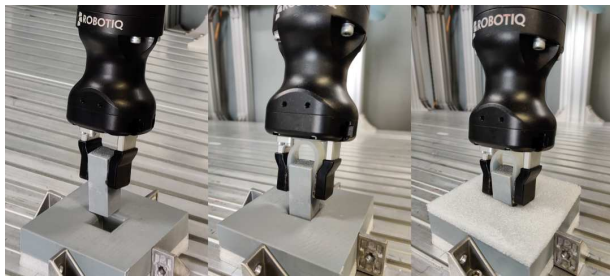}
\captionof{figure}{(left to right) High-, medium-, and low-stiffness environments.}
\label{fig:env-stiff}
\end{minipage}
\begin{minipage}[b]{.5\linewidth}
    \begin{tabular}{|c|c|c|c|}
    \hline
    \textbf{Method/Stiffness} & \textbf{High} & \textbf{Medium} & \textbf{Low} \\ \hline
    \textbf{Scratch}            & 100\%  & 70\%  & 40\% \\ \hline
    \textbf{Sim2real}           & 95\% & 100\% & 100\% \\ \hline
    \textbf{Ours}               & 100\% & 100\% & 100\% \\ \hline
    \end{tabular}
    \caption{Success rate of 3D-printed-cuboid insertion task with different degrees of contact stiffness.}
    \label{tab:stiffness-accuracy}
\end{minipage}\hfill
\end{table}

\subsubsection{\textit{Varying Insertion Tasks}} \label{subsubsec:varying-tasks}
\begin{figure}[ht]
    \centering
    \includegraphics[width=0.8\textwidth]{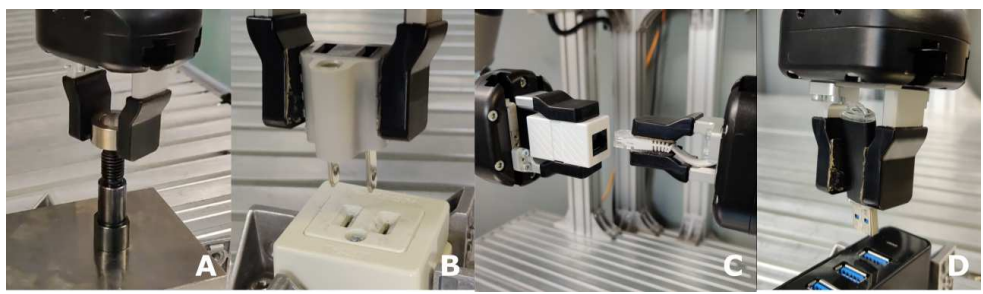}
    \caption{Several insertion tasks with different degrees of complexity. (A) Metal ring (high stiffness) with 0.2 mm of clearance. (B) Electric outlet requiring high insertion force. (C) Local-area-network (LAN) port, delicate with  complex shape. (D) Universal serial bus (USB).}
    \label{fig:tasks}
\end{figure}

    Lastly, we evaluate the learned policy on a series of novel insertion tasks, none seen during training, to assess its generalization capabilities. These insertion tasks included challenges such as adapting to a very hard surface (high stiffness), requiring a minimal insertion force to perform the insertion, and a complex peg shape for  mating  the parts. The different insertion scenarios are depicted in \Cref{fig:tasks}.

\begin{table}[ht]
\centering
\begin{tabular}{|c|c|c|}
\hline
\textbf{Task} & \textbf{Success rate} & \textbf{Insertion force} \\ \hline
\textbf{Ring} & 80\% & 5N \\ \hline
\textbf{Electric Outlet (x)} & 75\% & 10N \\ \hline
\textbf{Electric Outlet (y)} & 75\% & 10N \\ \hline
\textbf{LAN port (x)} & 55\% & 5N \\ \hline
\textbf{LAN port (y)} & 60\% & 5N \\ \hline
\textbf{USB} & 80\% & 8N \\ \hline
\end{tabular}
\caption{Success rate of learned policy on several insertion tasks.}
\label{tab:diff-tasks}
\end{table}

    For each task, the learned policy was executed 20 times from random initial positions and assuming perfect estimation of the goal position. \Cref{tab:diff-tasks} shows the success rate of the learned policy on these novel tasks, along with the desired insertion force set for each task. As the insertion force was defined as a policy input, we could define specific desired insertion force for each task. Even though the policy was only trained by using the simpler cuboid-peg insertion task, mainly in simulation and shortly refined on  a real robot with a 3D-printed version of the same task, the learned policy achieved a high success rate in novel and complex insertion tasks. 
    
    Compared to the cuboid-peg insertion task, on these novel insertion tasks, the peg was more likely to become stuck during the task's search phase, as the surrounding surface near the hole was not smooth and may have had  crevices. The extra challenges were not present during the training phase, which reduced the capability of the learned policy to react in an appropriate way. The insertion task of the LAN port was the most challenging for the policy due to the complex shape of the LAN cable endpoint. If just one corner of the LAN adapter was stuck, the insertion could not be completed even if large force was applied. 
    
    Additionally, we  tested the policy on different insertion planes for the electric outlet and the LAN port tasks. In both cases,  success rate was similar due to  training with the randomized insertion planes. However, the policy was slightly better with insertions on the y-axis plane due to  retraining (on the real robot) only being done  on this axis.

\subsection{Ablation Studies}\label{subsec:ablation}
    In this section, we evaluate the individual contribution of some components added to the proposed learning framework.
    
\subsubsection{Learning from Scratch vs Sim2real}
\begin{figure}[ht]
    \centering
    \includegraphics[width=0.5\textwidth]{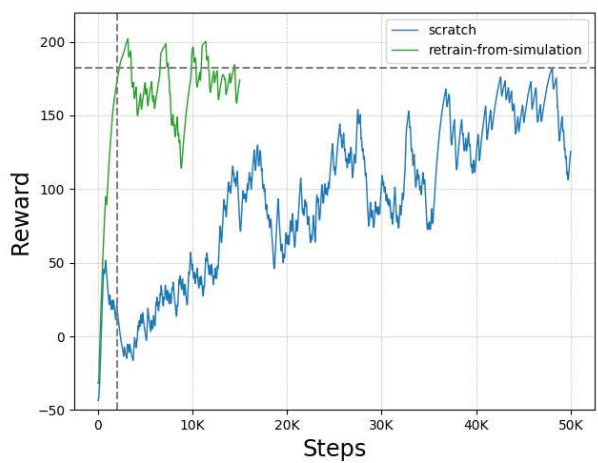}
    \caption{Comparison between learning from scratch and learning from a policy learned on simulation: learning curve for 3D-printed cuboid-peg insertion task on  real robot with random initial positions.}
    \label{fig:sim2real-comparison}
\end{figure}

    The inclusion of transfer learning from the simulation to the real robot for the proposed learning framework was evaluated. We compared the learning performance of training the agent on the real robot from scratch versus learning starting from a policy learned on simulation. Training from scratch was performed for 50,000 steps, while  retraining from the simulation lasted 15,000 steps. \Cref{fig:sim2real-comparison} shows the learning curve for both training sessions. Learning from scratch required at least 50,000 steps to succeed at the tasks most of the time. In contrast, learning from the pretrained policy on the simulation achieved the same performance in under 5000 steps. The policy from the simulation still required some training to fine-tune the controller to  real-world physics dynamics, which are difficult to simulate, as can be seen from the slow start and the drops in cumulative reward. 
    
\subsubsection{Policy Architecture}
\begin{figure}[ht]
    \centering
    \includegraphics[width=0.5\textwidth]{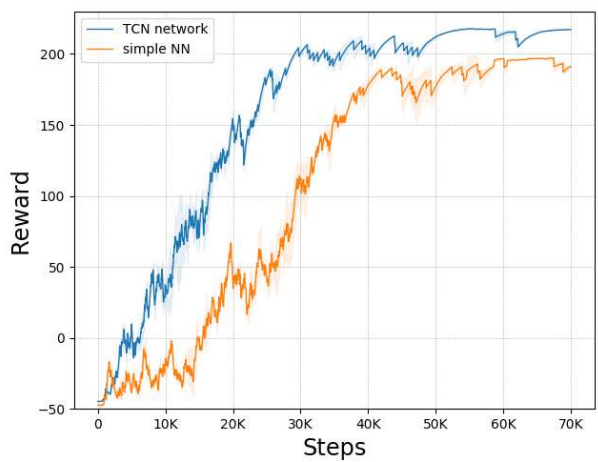}
    \caption{Comparison between policy architectures: learning curve for  cuboid-peg insertion task with random initial positions.}
    \label{fig:policy-arch-comparison}
\end{figure}

    We evaluated the contribution of the policy architecture introduced in our method (see \Cref{subsec:policy-arch}) by comparing it to a policy with a simple neural network (NN) with two fully connected layers as used in previous work \cite{beltranhern2020learning}. We trained both policies on the cuboid-peg insertion task on the simulation and compared their learning performance. \Cref{fig:policy-arch-comparison} shows the learning curve of both policy architectures for a training session of 70,000 time steps. From the figure, is clear that, with our newly proposed TCN-based policy, the agent was able to learn faster and exploit better rewards. The TCN-based policy learned a successful policy (25,000) about 15,000 steps faster than the simple neural-network (NN)-based policy did (40,000). Additionally, the TCN-based policy converged to a higher cumulative reward  than that of the simple NN-based policy. 

\subsubsection{Policy Inputs}
    Lastly, we evaluated the choice of inputs for the policy. We compared our proposed policy architecture with all inputs, as defined in \Cref{subsec:policy-arch}, with two variants. First, we considered the policy without the inclusion of  prior action $\textbf{a}_{t-1}$. Second, we considered the policy without  knowledge of  desired insertion force $F_g$. The training environment was the cuboid-peg insertion task on the simulation with a random initial position and random desired insertion force. In the case of the policy that did not have $F_g$ as input, the cost function still accounted for the desired insertion force. 
    
\begin{figure}[ht]
    \centering
    \includegraphics[width=0.5\textwidth]{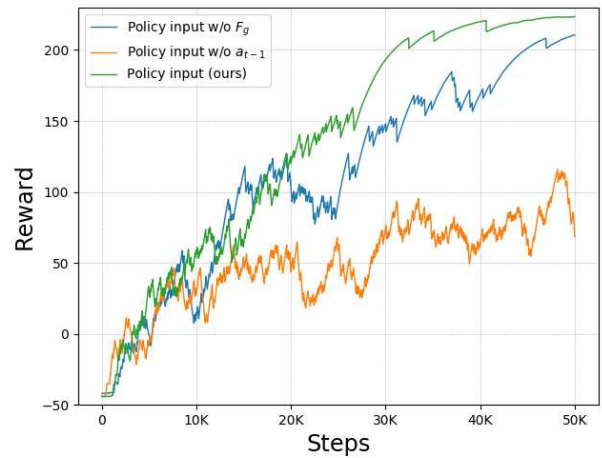}
    \caption{Comparison of policies with different inputs. Learning curve for  cuboid-peg insertion task with random initial positions and random desired insertion force.}
    \label{fig:policy-input}
\end{figure}
    
    \Cref{fig:policy-input} shows the comparison of the learning curves. Most notable is the poor performance of the policy that lacked the knowledge of  prior action $\textbf{a}_{t-1}$. Prior-action information is critical for the agent to more quickly converge to an optimal policy. Additionally,  knowledge of $F_g$ enables the agent to find policies that yield higher cumulative rewards, and to learn faster.

\section{Discussion}
    We  proposed a learning framework for position-controlled robot manipulators to solve contact-rich manipulation tasks. The proposed method allows for learning low-level high-dimensional control policies on real robotic systems. The effectiveness of the learned policies was shown through an extensive experiment study. We showed that the learned policies had a high success rate at performing the insertion task under the assumption of a perfect estimation of the goal position. The policy correctly learned the nominal trajectory and the appropriate force-control parameters to succeed at the task. The policy also achieved a high success rate under varying environmental conditions  in terms of uncertainty of  goal position, environmental stiffness, and novel insertion tasks.
    
    While   model free reinforcement-learning algorithm SAC was used in this work, the proposed framework can  easily be adapted to other RL algorithms. The choice of SAC was due to its sample efficiency as an off-policy algorithm. The pros and cons of using other learning algorithms would be  interesting future work.
    
    One limitation of our learning framework is the selection of the force-control parameter range (see \Cref{subsec:parallel-control}). The choice of a wide range of values may allow for the policy to adapt to very different environments, but it also increases the difficulty of learning a task, as small variations in the action may cause undesired behaviors, as was the case during the first 20,000 to 30,000 steps of training (see \Cref{fig:policy-arch-comparison}). On the other hand, a narrow range would make it easier and faster to learn a task, but it may not generalize well to different environments. Defining a range is much easier than manually finding the optimal parameters for each task, but it is still a manual process. Therefore, another interesting future study would be to use demonstrations to learn a rough estimation of the optimal force parameters to further reduce training times.

\reftitle{References}
\bibliography{main}





\vspace{6pt} 

\end{document}